\documentclass{article}
\usepackage{spconf,amsmath,graphicx}
\usepackage[hidelinks]{hyperref}
\usepackage[dvipsnames,table]{xcolor}
\usepackage{amsfonts}
\usepackage{enumitem} 
\usepackage{amsmath}
\usepackage{makecell}
\usepackage{booktabs}
\usepackage{graphicx}
\usepackage{enumitem}

\title{Mixture of Experts for Recognizing Depression from Interview and Reading Tasks}
\name{Loukas Ilias, Dimitris Askounis}
\address{DSS Laboratory, School of ECE, National Technical University of Athens, Greece}
\begin{document}
\ninept
\maketitle
\begin{abstract}
Depression is a mental disorder with psychological, physical, and social manifestations. Speech has been shown to be a reliable biomarker for early detection. However, existing methods typically (i) analyze only spontaneous speech overlooking informative cues from read speech; (ii) rely on transcripts which are often difficult to obtain (manual) or suffer from high word-error rates (automatic); and (iii) do not leverage input-conditional computation methods. To resolve these limitations, this is the first study in depression recognition task that jointly represents spontaneous and read speech, employs multimodal fusion, and integrates Mixture of Experts within a single end-to-end deep neural network. On the Androids corpus, our proposed approach achieves an Accuracy and F1-score of 87.00\% and 86.66\% respectively, demonstrating the value of combining speech styles with input-adaptive modeling.
\end{abstract}
\section{Introduction}

Depression comes with a series of emotional and physical problems hindering the person's ability to carry out daily activities \cite{kanter2008nature}. According to the World Health Organization\footnote{https://www.who.int/news-room/fact-sheets/detail/depression}, approximately 3.8\% of the population suffers from depression, with suicide constituting the fourth leading cause of death in ages between 15 and 29 years old. The study in \cite{CUMMINS201510} shows that depression entails both cognitive and physiological changes affecting speech production processes. Therefore, speech constitutes a reliable biomarker for recognizing depression. Early recognition of depression is crucial for ensuring a better quality of life.

Existing studies \cite{ilias24_interspeech} use multimodal methods by combining transcripts and audio files. However, manual transcripts are not always available and thus automatic speech recognition systems are used. These systems often have high word-error rate in languages other than english and thus insert errors in machine learning models. Additionally, literature suggests that both read and spontaneous speech should be considered by clinical depression collections in their protocols \cite{7178876}. However, the majority of the existing studies utilize only spontaneous speech tasks, i.e., description of picture, interview, with only a few studies incorporating read speech \cite{6639130, 10822855}. Traditional approaches obtain representation vectors of the input modalities and then use the same single layer for the batch of the representation vectors. On the other hand, Mixture of Experts (MoEs) have emerged as a powerful and effective approach for increasing evaluation performance by employing some experts.

In order to tackle these limitations, we present the first study, which uses both spontaneous and read speech, multimodal fusion methods, and MoE models into a single end-to-end trainable deep neural network (DNN). Specifically, we convert audio files of read and spontaneous speech into images of three channels, namely log-Mel spectrogram, velocity, acceleration. We pass these images through two shared AlexNet pretrained models (one model for each input). Next, the outputs of the AlexNet model are used as input to a multimodal fusion approach, namely BLOCK \cite{Ben-younes_Cadene_Thome_Cord_2019}, which generalizes both concepts of rank and mode ranks for tensors. Finally, we use three variants of MoE models, namely sparse MoEs \cite{shazeer2017} and two variants of multilinear MoEs based on factorization \cite{oldfield2024mumoe}. Experiments conducted on Androids corpus \cite{tao23_interspeech} show that our proposed approach obtains valuable advantages over state-of-the-art approaches.

Our main contributions can be summarized as follows:
\begin{itemize}
    \item We present a new method to recognize depression from read and spontaneous speech.
    \item We employ three variants of MoE layers and compare their performances.
    \item We perform a series of ablation experiments to verify the effectiveness of the introduced approach.
\end{itemize}

\section{Related Works}

\noindent \textbf{Depression Recognition from Speech.} The authors in \cite{ilias24_interspeech} introduced a DNN consisting of a cross-attention layer and multimodal fusion methods. They used both speech and transcripts as inputs to the proposed DNN. The authors investigated via a multi-task learning (MTL) setting if gender, age, and education level improve depression recognition performance. In \cite{tao20_interspeech}, the authors extracted a set of features from read speech, including energy (loudness), Mel-Frequency Cepstral Coefficients (MFCC), F0, and more, and trained a Support Vector Machine (SVM) classifier. The study in \cite{6639130} utilized both read and spontaneous speech samples to recognize depression. A feature extraction approach was adopted by the authors followed by the train of an SVM classifier. Findings showed that spontaneous speech gave better results than read speech. Jitter, shimmer, loudness, and energy were proved robust features. The authors in \cite{huang2024depression} utilized the DAIC-WOZ dataset and fine-tuned wav2vec2 pretrained model for recognizing depression. Gupta et al. \cite{GUPTA2025101710} presented a multitask learning framework to recognize major depressive disorder and post-traumatic stress disorder (PTSD). In terms of the architecture, the authors used Mel-spectrograms and passed them through CNN layers followed by LSTMs. Findings suggested that MTL performed better than the single-task learning framework. The authors in \cite{10448231} presented a mutual information based approach to recognize depression. The aim of the study was to maximize depression information, while minimizing at the same time speaker information. Results demonstrated the effectiveness of the proposed approach. The study in \cite{DAS2024105898} extracted both MFCCs and spectrogram from audio files and trained a deep neural network based on CNNs for recognizing depression. Linear predictive coding and MFCC features were extracted in \cite{DU2023299}. Next, the authors trained a deep neural network consisting of CNNs and LSTMs. Results demonstrated the effectiveness of both production and perception features in depression recognition task.  Phoneme-based features were used in \cite{MUZAMMEL2020100005}. Specifically, the authors used spectrograms of vowels and consonants as inputs to CNN models. Finally, a deep neural network based on the fusion of these models was trained. Results showed that the fusion of both networks yielded the highest evaluation results. A set of features, including F0, jitter, shimmer, loudness, MFCC, voicing probability, and more, were extracted by \cite{10.3389/fnbot.2021.684037}. The authors trained a deep learning model consisting of LSTMs and Multihead Attention layer. 

\noindent \textbf{Mixture of Experts.} The idea of Mixture of Experts was originally proposed in \cite{6797059} and is based on the divide-and-conquer approach. Instead of using a same single layer for the inputs, MoE models consist of expert layers and a routing (or gating) network. The expert layers are usually simply dense layers, while the routing network is responsible for determining which experts can be used for the input. Then, the outputs of each expert are aggregated through a weighted average. Multiple levels of hierarchy are also employed \cite{716791}. Many variations of MoE models have been proposed throughout the years. In \cite{shazeer2017}, the authors introduced the sparsely-gated MoE layer, which computes a weighted sum of the outputs from only the top-\textit{k} experts, rather than aggregating the outputs from all the experts. However, sparse MoEs have the limitations of training instability, parameter-inefficiencies, and non-differentiable nature \cite{puigcerver2024from}. To tackle these limitations, a recent study \cite{oldfield2024mumoe} introduces Multilinear MoE layers, namely $\mu MoE$. MoE models have found extensive applications in several domains, including natural language processing, speech processing, computer vision \cite{fedus2022review, cai2024survey}, seizure subtype classification \cite{10335740}, patient heterogeneity (electronic health records in intensive care units) \cite{9508549}. 
\section{Dataset}

We use the Androids Corpus \cite{tao23_interspeech} to conduct our experiments. The Androids corpus consists of a reading and an interview task. We use both tasks in this study. In terms of the interview task, the person is asked to answer to some questions about everyday life. Interview task corresponds to a spontaneous speech task. Regarding the reading task, persons are asked to read a short fairy tale written by Aesop, namely "The Wind of the North and the Sun". Androids corpus includes 52 participants in the control group and 58 people in the depression group. The authors of \cite{tao23_interspeech} have conducted a chi-squared test, which reveals that there is no difference between the control group and the depression one in terms of gender and education level. Similarly, results of a two-tailed t-test demonstrated that there is no difference in terms of age distribution. Therefore, speech differences are attributable only to the speech pathologies.

\section{Methodology}

\begin{figure*}[!htb]
    \centering
    \includegraphics[width=0.8\linewidth]{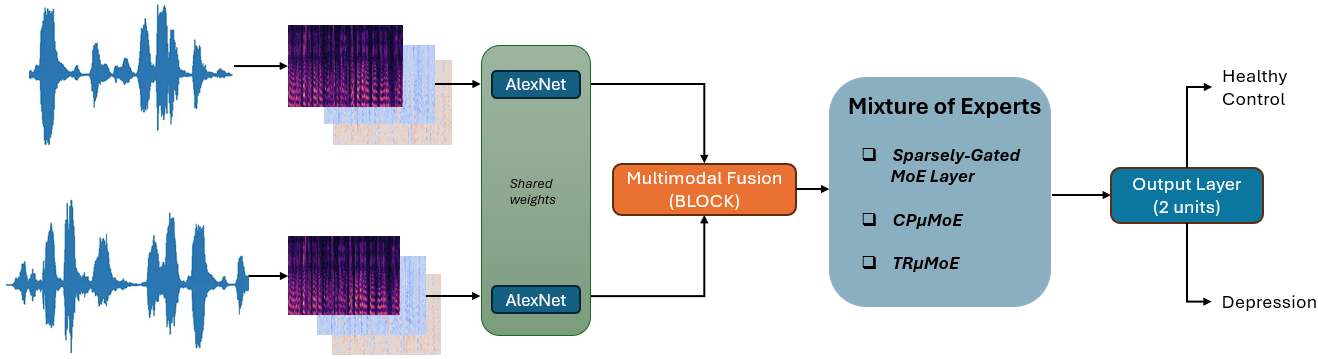}
    \caption{Our proposed Methodology. }
    \label{methodology}
\end{figure*}

In this section, we describe our proposed methodology for recognizing depression through read and spontaneous speech. In Fig.~\ref{methodology}, our proposed methodology is illustrated.

\noindent \textbf{Reading Task.} Each audio file is converted into an image of three channels, namely log-Mel spectrogram, its velocity ($\Delta$), and its acceleration ($\Delta\Delta$). Specifically, we use \textit{librosa} \cite{mcfee2015librosa}. For obtaining the log-Mel spectrogram, we use 224 Mel bands, hop length accounting for 512, and a Hanning window. Each image is resized to $(224 \times 224)$ pixels. Let $f^{read}$ be the image corresponding to the reading task.

\noindent \textbf{Interview Task.} We adopt the same methodology to obtain an image per audio file. Let $f^{interview}$ be the image corresponding to the interview task.

\noindent \textbf{Deep Neural Network.} Both $f^{read}$ and $f^{interview}$ are passed through two pretrained AlexNet models sharing the same weights \cite{krizhevsky2014one}. We modify the last layer of the AlexNet model, so as to ensure that the output of the AlexNet model has a dimensionality of 768d. Let $f^{interview} _{AlexNet}, f^{read} _{AlexNet} \in \mathbb{R}^{768}$ denote the output of the AlexNet pretrained model corresponding to the interview and reading task respectively.
    
\noindent \textbf{BLOCK fusion.} We pass both $f^{interview} _{AlexNet}$ and $f^{read} _{AlexNet}$ through the BLOCK fusion introduced in \cite{Ben-younes_Cadene_Thome_Cord_2019}. Specifically, this multimodal fusion method is based on the block-term tensor decomposition \cite{doi:10.1137/070690729} \cite{carroll1970analysis}. Let the output of this component be $f_{fusion} \in \mathbb{R}^d$, where $d=768$.
    
\noindent \textbf{Mixture of Experts -- Output.} We pass $f_{fusion}$ through MoE models, which are described in detail below:

\begin{itemize}[leftmargin=*]
	\setlength{\itemsep}{0pt}    
\item \textbf{Sparse MoE}: The output of Sparse MoE is denoted as $y$ and is given by: $y = \sum_{i=1}^n G(x)_i E_i (x)$. Let $G(x)_i$ denote the gating (router) network, while $E_i (x)$ indicates the output of the $ith$ expert. Specifically, $G(x)_i$ indicates the weight assigned to each expert by the gating network. $n$ denotes the number of experts. In our experiments, we use two-layer MLPs as expert networks. 

Next, we describe the method for obtaining the gating coefficients.

we add noise to the input ($H(x)_i = (x \cdot W_g)_i + StandardNormal() \cdot Softplus \left(\left(x \cdot W_{noise} \right)_i \right)$), where $StandardNormal()$ indicates standard normal distribution. Next, we keep only the top-\textit{k} values (see Eq.~\ref{keep_topk}). Adding noise facilitates load balancing among the experts, while keeping only the top-\textit{k} most relevant experts saves computation, since only a few of the experts are activated via a gating network.

\begin{equation} 
        \resizebox{\linewidth}{!}{$
        \text{KeepTopK}(v, k)_i = \begin{cases} v_i & \text{if } v_i \text{ is in the top } k \text{ elements of } v, \\ 
            -\infty & \text{otherwise.}
            \end{cases}$}
            \label{keep_topk}
        \end{equation}

Finally, we apply the softmax activation function to get the coefficients  $G(x) = Softmax \left(KeepTopK\left(H\left(x\right),k\right)\right) $.

Next, we describe the \textbf{loss function}, which is minimized in this study. 
\begin{itemize}[leftmargin=*]
    \item $L_{imp}$: This loss is used for ensuring uniform gating weights for all experts. In this way, the phenomenon of producing large weights to specific experts is avoided.

This loss is equal to the square of the coefficient of variation of the set of importance values and is defined via:$
            L_{\text{Imp}}(\mathcal{B}) = \text{CV} \left( \left\{ \text{Imp}_i(\mathcal{X}) \right\}_{i=1}^n \right)^2
            $, 
        where 
        $
            \text{CV}(\cdot) = \frac{\text{Std}(\cdot)}{\text{Mean}(\cdot)},$ and $
            \text{Imp}_i(\mathcal{X}) = \sum_{x \in \mathcal{X}} G(x)
            $.
\item         $L_{Load}$: This loss function ensures balanced loads across all experts, i.e., each expert receives equal number of training samples. 
        $
            L_{\text{Load}}(\mathcal{B}) = \text{CV} \left( \left\{ \text{Load}_i(\mathcal{X}) \right\}_{i=1}^n \right)^2
        $, where $            \text{Load}_i(\mathcal{X}) = \sum_{x \in \mathcal{X}} P_i(x)$ . $Load_i$ denotes the number of training examples per expert and $P_i(x)$ denotes the probability that $G(x)_i$ is non-zero. We define $P_i(x)$ as: $P_i(x) = \Phi \left ( \frac{\left(xW_G \right)_i - kth\_excluding \left(H (x),k, i \right)}{Softplus \left(\left (xW_{noise}\right)_i \right)} \right)$, \\ where $kth\_excluding (H (x), k, i )$ is the $k$th highest component of $H$ excluding the $i$th component. $\Phi$ is the cumulative distribution function of the standard normal distribution.
\item Loss Function: $L = L_{cross\_entropy} + \alpha \cdot (L_{imp} + L_{load})$,
        where $\alpha$ is a hyperparameter.

\end{itemize}            

\item \textbf{CP$\mu$MoE \cite{oldfield2024mumoe}.} This method is based on the multilinear MoE layer ($\mu MoE$). A $\mu MoE$ layer consists of $N$ experts and is parameterized by a weight tensor $W \in \mathbb{R}^{N \times I \times O}$ and expert gating parameter $G \in \mathbb{R}^{I \times N}$, where $I$ and $O$ indicate the input and output dimensions respectively. A $\mu MoE$ layer takes a sparse, convex combination of $N$ explicit experts’ latent factors. CP$\mu$MoE relies on a CP decomposition \cite{carroll1970analysis, hitchcock1927expression} of the weight tensor. CP$\mu$MoE reduces the parameters from $NIO$ to $R(N+I+O)$, where $R$ denotes the rank. We minimize the cross-entropy loss function.
        
\item \textbf{TR$\mu$MoE \cite{oldfield2024mumoe}}: Similar to CP$\mu$MoE, this method is based on the $\mu MoE$ layer. TR$\mu$MoE relies on the Tensor Ring Decomposition \cite{zhao2016tensor} and reduces the parameters to $ (R1NR2 + R2IR3 + R3OR1)$. We minimize the cross-entropy loss function.

\end{itemize}

\section{Experiments and Results}

\noindent \textbf{Baselines}
\begin{itemize}[leftmargin=*]
	\setlength{\itemsep}{0pt}    
    \item \textit{Silences \cite{tao20_interspeech}}: This method extracts a set of features, including the number and total length of silences, and trains an SVM classifier.
    \item \textit{Only speech \cite{ilias24_interspeech}}: This method converts audio into log-Mel spectrogram, delta, and delta-delta and finetunes a pretrained AlexNet model.
    \item \textit{BS1 \cite{tao23_interspeech}}: Features per window frame are extracted and used to train an SVM classifier.
    \item \textit{BS2 \cite{tao23_interspeech}}: Features per window frame are extracted and used to train an LSTM layer, while the final prediction is obtained via majority vote.
\end{itemize}

\begin{table}[htbp]
\scriptsize
\centering
\caption{Performance comparison among proposed models and baselines. Best results per evaluation metric are in bold.}
\begin{tabular}{lccccc}
\toprule
\multicolumn{1}{l}{}&\multicolumn{5}{c}{\textbf{Evaluation metrics}}\\
\cline{2-6} 
\multicolumn{1}{l}{\textbf{Architecture}}&\textbf{Precision}&\textbf{Recall}&\textbf{F1-score}&\textbf{Accuracy}&\textbf{Specificity}\\
\midrule
\multicolumn{6}{>{\columncolor[gray]{.8}}l}{\textbf{Comparison with state-of-the-art}} \\
\textit{Silences \cite{tao20_interspeech}} & 84.50 & 84.60 & 84.55 & 84.50 & - \\ \hline
\textit{Only speech \cite{ilias24_interspeech}} & 80.73 & 85.70 & 82.49 & 80.52 & 74.21 \\ \hline
\textit{BS1 \cite{tao23_interspeech}} & 73.50 & 74.50 & 73.60 & 73.30 & - \\ \hline
\textit{BS2 \cite{tao23_interspeech}} & 85.80 & 86.10 & 84.70 & 83.90 & - \\ 
\midrule
\multicolumn{6}{>{\columncolor[gray]{.8}}l}{\textbf{Introduced Approaches}} \\
\textit{Sparse MoE} & 84.05 & 84.91 & 83.92 & 83.87 & 81.10 \\ & $\pm$11.81 & $\pm$8.37 & $\pm$7.82 & $\pm$7.76 & $\pm$13.22 \\ \hline
\textit{CP$\mu$MoE} & 85.81 & 83.63 & 84.23 & 85.25 & \textbf{84.81} \\ 
& $\pm$10.79 & $\pm$12.47 & $\pm$9.82 & $\pm$8.80 & $\pm$10.51 \\ \hline
\textit{TR$\mu$MoE} & \textbf{86.80} & \textbf{87.10} & \textbf{86.66} & \textbf{87.00} & \textbf{84.81} \\
& $\pm$9.02 & $\pm$8.99 & $\pm$7.44 & $\pm$6.64 & $\pm$10.51 \\

\bottomrule
\end{tabular}
\label{performance_comparison}
\end{table}

\noindent \textbf{Experimental Setup.} In terms of the MoE layer, we use 4 experts and keep the 3 most relevant ones. We set $\alpha$ equal to 0.1. Regarding $CP\mu MoE$ and $TR\mu MoE$, we set: $I=768$, $O=128$, and $N=3$. In terms of $CP\mu MoE$, we set $R=4$. With regards to $TR\mu MoE$, we set $R1=R2=R3=4$. We use a learning rate of \texttt{1e-4}. We use the Adam optimizer. We train our proposed models for 30 epochs with a batch size of 8. Experiments are conducted on a 5-fold cross-validation setting. Experiments are ran four times. Experiments are conducted on a NVIDIA A100 80GB PCIe GPU.

\noindent \textbf{Evaluation Metrics.} Accuracy, Precision, Recall, and F1-score have been used to evaluate the results of our proposed approach. We report the mean and standard deviation of these metrics over four runs.

\noindent \textbf{Results.} Results of our proposed methodology are reported in Table~\ref{performance_comparison}. As one can observe, $TR\mu MoE$ is our best performing model outperforming the rest of our introduced approaches in Accuracy by 1.75-3.13\%, in Recall by 2.19-3.47\%, in Precision by 0.99-2.75\%, and in F1-score by 2.43-2.74\%. Differences in performance between $TR\mu MoE$ and $CP\mu MoE$ are attributable to the factorization method used. Specifically, it is shown that Tensor Ring Factorization is a more powerful method than CP decomposition in our task. We also observe that Sparsely-Gated MoE layer presents lower evaluation results than both $TR\mu MoE$ and $CP\mu MoE$. We speculate that this difference is attributable to the inherent limitations of sparse MoE layers, including training instability, non-differentiable nature, and parameter-inefficiency. Values of standard deviations are in alignment with existing literature \cite{escobargrisales23_interspeech,laquatra24_interspeech} and are attributable to the limited datasets used. In comparison with baselines, we observe that our best performing model surpasses these approaches in terms of Accuracy by 2.50-13.70\%, Recall by 1.00-12.60\%, F1-score by 1.96-13.06\%, and Precision by 1.00-13.30\%. These differences demonstrate the advantages of combining multimodal fusion methods and MoE layers into a single DNN.

\begin{table}[htbp]
\scriptsize
\centering
\caption{Ablation Study. Best results per evaluation metric are in bold.}
\begin{tabular}{lccccc}
\toprule
\multicolumn{1}{l}{}&\multicolumn{5}{c}{\textbf{Evaluation metrics}}\\
\cline{2-6} 
\multicolumn{1}{l}{\textbf{Architecture}}&\textbf{Precision}&\textbf{Recall}&\textbf{F1-score}&\textbf{Accuracy}&\textbf{Specificity}\\
\midrule
\textit{read} & 79.06 & 79.98 & 78.87 & 79.41 & 78.82 \\ 
& $\pm$13.85 & $\pm$13.91 & $\pm$12.15 & $\pm$10.97 & $\pm$14.97 \\ \hline
\textit{spontaneous} & 81.25 & 84.72 & 82.14 & 81.73 & 77.68 \\ & $\pm$13.04 & $\pm$8.61 & $\pm$8.16 & $\pm$8.99 & $\pm$17.11 \\ \hline
\textit{Non-shared} & 84.50 & 86.81 & 84.93 & 84.71 & 80.29 \\ & $\pm$11.99 & $\pm$11.85 & $\pm$9.98 & $\pm$9.90 & $\pm$17.67 \\ \hline
\textit{Concatenation} & 86.51 & 83.84 & 84.38 & 85.08 & \textbf{86.12} \\ 
& $\pm$12.40 & $\pm$11.78 & $\pm$9.53 & $\pm$8.73 & $\pm$12.98 \\ \hline
\textit{-- MoE layer} & 84.81 & 84.80 & 84.03 & 83.69 & 81.98 \\ 
& $\pm$11.84 & $\pm$9.24 & $\pm$7.55 & $\pm$7.89 & $\pm$16.12 \\ \hline
\textit{\textbf{Methodology}} & \textbf{86.80} & \textbf{87.10} & \textbf{86.66} & \textbf{87.00} & 84.81 \\
& $\pm$9.02 & $\pm$8.99 & $\pm$7.44 & $\pm$6.64 & $\pm$10.51 \\
\bottomrule
\end{tabular}
\label{ablation_study}
\end{table}

\begin{table}[!htbp]
\scriptsize
\centering
\caption{Ablation Study (Fusion). Best results per evaluation metric are in bold.}
\begin{tabular}{lccccc}
\toprule
\multicolumn{1}{l}{}&\multicolumn{5}{c}{\textbf{Evaluation metrics}}\\
\cline{2-6} 
\multicolumn{1}{l}{\textbf{Architecture}}&\textbf{Precision}&\textbf{Recall}&\textbf{F1-score}&\textbf{Accuracy}&\textbf{Specificity}\\
\midrule
\multicolumn{6}{>{\columncolor[gray]{.8}}l}{\textbf{Ablation Experiments}} \\
\textit{GMU} & 81.07 & 80.19 & 79.94 & 80.61 & 80.42 \\ 
& $\pm$10.48 & $\pm$12.56 & $\pm$9.72 & $\pm$7.84 & $\pm$12.01 \\ \hline
\textit{MUTAN} & 85.21 & 83.22 & 83.75 & 84.62 & 84.38 \\ 
& $\pm$12.50 & $\pm$13.73 & $\pm$11.86 & $\pm$10.85 & $\pm$13.92 \\ \hline
\textit{MLB} & 85.53 & 81.05 & 81.82 & 83.65 & 82.90 \\ 
& $\pm$11.90 & $\pm$17.50 & $\pm$13.71 & $\pm$10.49 & $\pm$13.17 \\ \hline
\textit{MFB} & 85.57 & 84.77 & 84.67 & 85.83 & \textbf{85.28} \\ 
& $\pm$10.12 & $\pm$15.09 & $\pm$11.48 & $\pm$10.00 & $\pm$9.42 \\ \hline
\textit{MFH} & 83.74 & 80.66 & 81.45 & 82.78 & 81.79 \\ 
& $\pm$10.99 & $\pm$14.85 & $\pm$11.46 & $\pm$9.86 & $\pm$13.89 \\ \midrule
\multicolumn{6}{>{\columncolor[gray]{.8}}l}{\textbf{Proposed Methodology}} \\
\textit{\textbf{Methodology}} & \textbf{86.80} & \textbf{87.10} & \textbf{86.66} & \textbf{87.00} & 84.81 \\
& $\pm$9.02 & $\pm$8.99 & $\pm$7.44 & $\pm$6.64 & $\pm$10.51 \\
\bottomrule
\end{tabular}
\label{ablation_fusion}
\end{table}

\noindent \textbf{Ablation Study.} In this section, we conduct a series of ablation experiments to prove the effectiveness of the proposed approach. Results are reported in Table~\ref{ablation_study}. Firstly, we use as input only the read speech and thus remove the multimodal fusion component. Results state that \textit{Only Read Speech} yileds an Accuracy of 79.41\%, which corresponds to a decline of 7.59\% in comparison with our proposed framework. Secondly, we use only spontaneous speech. Findings show that an Accuracy and F1-score of 81.73\% and 82.14\% respectively are obtained. Thirdly, we use two AlexNet models without shared weights. Results show that an Accuracy and F1-score of 84.71\% and 84.93\% respectively are obtained. Thus, fine-tuning two pretrained AlexNet models is a complex task for our limited dataset. Next, we concatenate the representations obtained through read and spontaneous speech instead of using BLOCK fusion. A decline in Accuracy and F1-score by 1.92\% and 2.28\% respectively is observed compared to our introduced methodology. Next, we remove the MoE component and use a dense layer of 128 units. Results show that Accuracy and F1-score drop to 83.69\% and 84.03\% respectively.

Additionally, we assess the usage of fusion methods. Results are reported in Table~\ref{ablation_fusion}. Instead of BLOCK, we use Gated Multimodal Unit (GMU) \cite{arevalo2020gated}, MUTAN decomposition \cite{8237547}, Multimodal Low-rank bilinear (MLB) pooling \cite{kim2017hadamard}, Multimodal factorized bilinear (MFB) pooling \cite{8334194}, and Multimodal factorized high-order (MFH) pooling \cite{8334194}. In our results, GMU achieved the lowest F1-score of 79.94\%, indicating its limited capacity for modeling complex cross-modal dynamics. MUTAN, which relies on Tucker decomposition to enable efficient bilinear interactions, performs moderately better with an F1-score of 83.75\%, though it still falls short in recall and accuracy compared to higher-order methods. MLB, which compresses bilinear interactions via low-rank approximations, reaches an F1-score of 81.82\%, also demonstrating suboptimal recall. MFB and MFH incorporate factorized bilinear pooling mechanisms that model multiplicative interactions more effectively; MFB achieves a relatively high F1-score of 84.67\%, and MFH adds further capacity by modeling higher-order interactions, although its F1-score drops slightly to 81.45\%. In contrast, our proposed fusion strategy using block-term tensor decomposition is designed to harness the benefits of both low-rank structure and high-order interactions and achieves the highest F1-score of 86.66\%, which is an improvement of 6.72\% over GMU, 2.91\% over MUTAN, 4.84\% over MLB, 1.99\% over MFB, and 5.21\% over MFH. Compared to the best-performing baseline (MFB), this represents a 1.23\% increase in precision and a 2.33\% increase in recall. In terms of accuracy, our method reaches 87.00\%, outperforming GMU by 6.39\%, MUTAN by 2.38\%, MLB by 3.35\%, MFB by 1.17\%, and MFH by 4.22\%. 

\begin{figure}[!htb]
    \centering
    \includegraphics[width=0.75\linewidth]{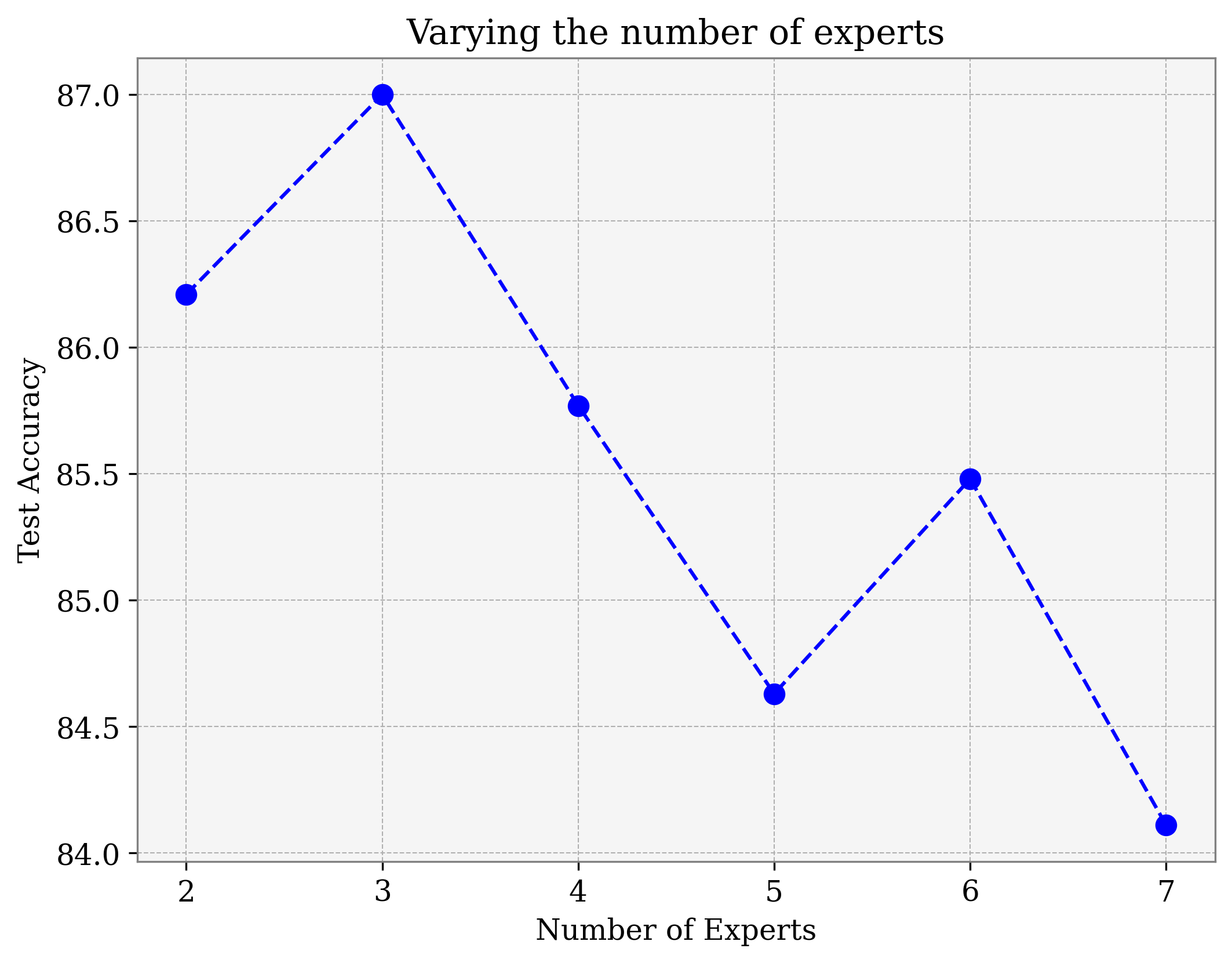}
    \caption{Test accuracy with respect to the number of experts}
    \label{ablation_study_number_of_experts}
\end{figure}

Finally, we modify the number of experts in terms of the MoE layer. Results are presented in Fig.~\ref{ablation_study_number_of_experts}. We observe that as the number of experts increases, accuracy decreases. We speculate that this difference in performance is attributable to the limited dataset used.

\section{Conclusion}

In this paper, we present the first study utilizing both spontaneous and read speech in the Italian language, multimodal fusion methods, and MoE models (sparse, mulilinear based on factorization) in a single neural network. Results show that multilinear MoE layers based on Tensor Ring Decomposition yielded the highest performance reaching Accuracy and F1-score up to 87\% and 86.66\% respectively. Results of an ablation study verified the effectiveness of the proposed approach. \textbf{Limitations:} We used one limited dataset consisting of 110 samples. Additionally, our approach depends on labelled datasets. However, obtaining large labelled datasets in healthcare domain is a challenging task due to privacy issues. \textbf{Future Work:} In the future, we aim to use self-supervised learning approaches and parameter-efficient fine-tuning strategies in conjunction with MoE variants.

\bibliographystyle{IEEEbib}
\bibliography{strings,refs}

\end{document}